\pdfoutput=1

\documentclass[11pt]{article}

\usepackage[]{ACL2023}

\usepackage{times}
\usepackage{latexsym}

\usepackage[T1]{fontenc}

\usepackage[utf8]{inputenc}

\usepackage{microtype}

\usepackage{inconsolata}

\usepackage{colortbl}  
\usepackage{xcolor}
\usepackage{array} 

\usepackage{tabularx}

\usepackage{multirow}

\usepackage{times}
\usepackage{url}
\usepackage{latexsym}
\usepackage{graphicx}
\usepackage{booktabs}
\usepackage{indentfirst}
\usepackage{pdfpages}

\title{DISC-FinLLM: A Chinese Financial Large Language Model \\ based on Multiple Experts Fine-tuning}

\author{
    Wei Chen\textsuperscript{\rm 1,2}\thanks{~~Wei Chen, Qiushi Wang, Zefei Long, Xianyin Zhang contribute equally to this work.}, 
    Qiushi Wang\textsuperscript{\rm 1},
    Zefei Long\textsuperscript{\rm 1}, 
    Xianyin Zhang\textsuperscript{\rm 1}, \\
    \textbf{Zhongtian Lu\textsuperscript{\rm 1}, 
    Bingxuan Li\textsuperscript{\rm 1},  
    Siyuan Wang\textsuperscript{\rm 1}, } \\ 
    \textbf{Jiarong Xu\textsuperscript{\rm 3}, 
    Xiang Bai\textsuperscript{\rm 2},
    Xuanjing Huang\textsuperscript{\rm 4},
    Zhongyu Wei\textsuperscript{\rm 1,\rm 5}\thanks{~~Corresponding Author.}}
    \\ [.3cm]
    \textsuperscript{\rm 1}{School of Data Science, Fudan University, China} \\
    \textsuperscript{\rm 2}{School of Software Engineering, Huazhong University of Science and Technology, China} \\
    \textsuperscript{\rm 3}{School of Management, Fudan University, China} \\   
    \textsuperscript{\rm 4}{School of Computer Science, Fudan University, China} \\        
    \textsuperscript{\rm 5}Research Institute of Intelligent Complex Systems, Fudan University, China \\    
    \{lemuria\_chen,xbai\}@hust.edu.cn \\
    \{qswang23,zflong23,xianyinzhang22,ztlu22,bxli16\}@m.fudan.edu.cn \\
    \{sywang18,jiarongxu,xjhuang,zywei\}@fudan.edu.cn 
}

\begin{document}
\maketitle


\begin{abstract}

We propose Multiple Experts Fine-tuning Framework to build a financial large language model (LLM), DISC-FinLLM. Our methodology improves general LLMs by endowing them with multi-turn question answering abilities, domain text processing capabilities, mathematical computation skills, and retrieval-enhanced generation capabilities. We build a financial instruction-tuning dataset named DISC-FIN-SFT, including instruction samples of four categories (consulting, NLP tasks, computing and retrieval-augmented generation). Evaluations conducted on multiple benchmarks demonstrate that our model performs better than baseline models in various financial scenarios. Further resources can be found at \emph{\url{https://github.com/FudanDISC/DISC-FinLLM}}. 


\end{abstract}

\section{Introduction}







The financial industry presents unique challenges and opportunities for Natural Language Processing (NLP) models~\cite{huang2020deep}. Traditional financial NLP models have made progress in various tasks such as news sentiment analysis~\cite{araci2019finbert}, financial event extraction~\cite{zheng2019doc2edag,yang2019using}, financial report generation~\cite{chapman2022towards}, stock price prediction~\cite{chen2018incorporating} and financial text summarization~\cite{la2020end}. However, as the quantity and complexity of financial data continue to increase, traditional financial NLP models face several limitations. These limitations include scarcity of human labeled data, insufficient financial specific knowledge, lack of multitasking capabilities, lack of ability in numerical computation, inability to handle real-time information, etc~\cite{gupta2020comprehensive}. Therefore, a comprehensive intelligent solution to effectively handle various tasks in the financial domain is still demanding.






Recently, the emergence of powerful commercial large language models (LLMs), like ChatGPT~\cite{openai2023chatgpt} and GPT-4~\cite{openai2023gpt4}, has unlocked the potential for innovation in financial artificial intelligence~\cite{zaremba2023chatgpt}. These models have impressed with their robust language understanding, dialogue skills and ability to follow instructions~\cite{ouyang2022training}. However, LLMs in general domain usually lack comprehensive knowledge of financial domain, especially in Chinese market. Thus, it becomes imperative to develop an open-source Chinese financial LLM that can support various user groups in different scenarios.

\begin{figure*}
      \centering
      \includegraphics[width=\textwidth]{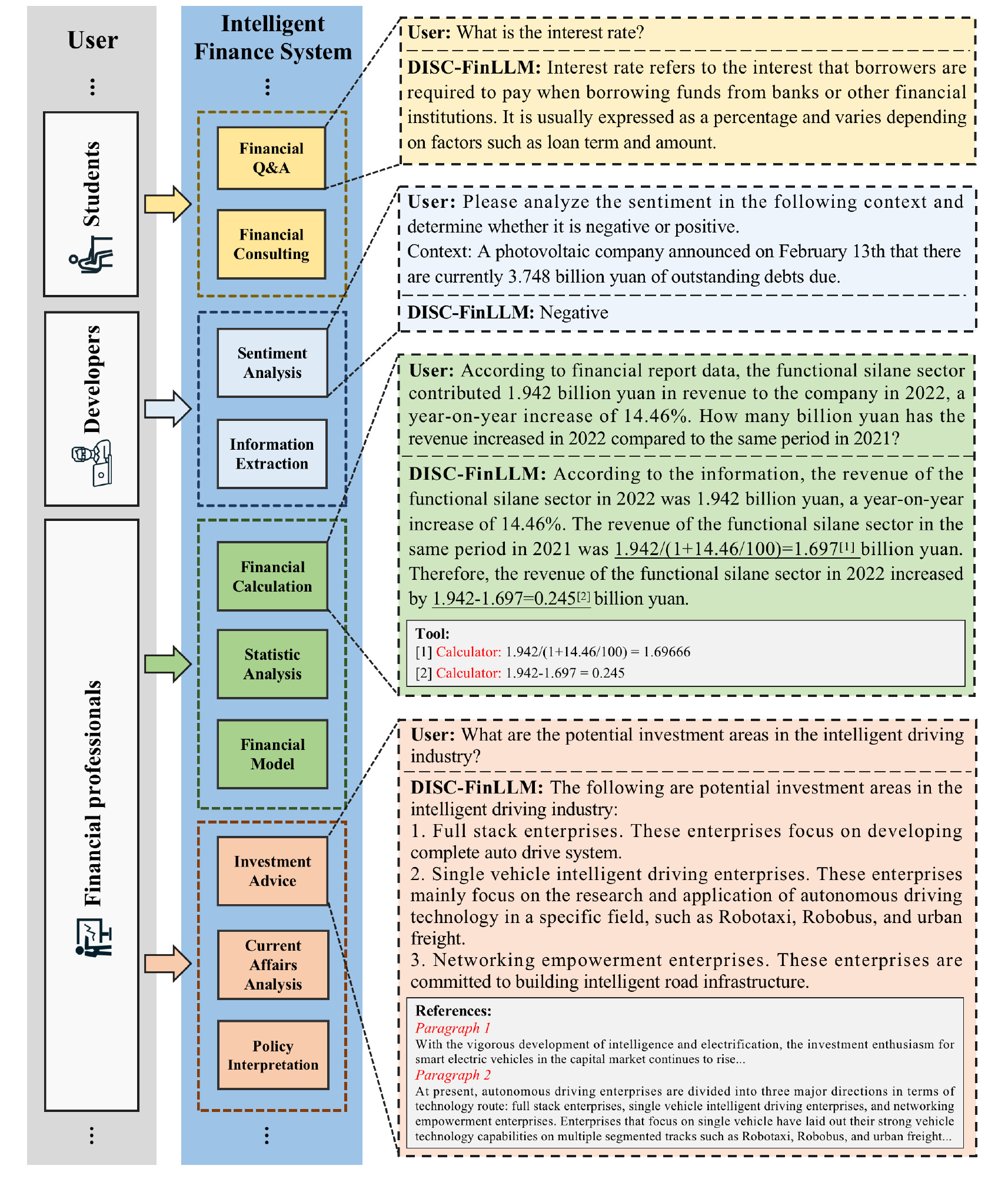}
      \caption{Overview of DISC-FinLLM serving different user groups in various financial scenarios.}
      \label{fig:MEFF}
\end{figure*}

In this paper, we propose a comprehensive approach to build Chinese financial LLMs and present DISC-FinLLM. Our method aims to enhance general LLMs by equipping them with the skills to address typical needs for financial text generation and understanding, meaningful multi-turn conversations on financial topics, and plugin functionality to support financial modeling and knowledge-enhanced system. To achieve these objectives, we create a rich supervised instruction dataset called DISC-FIN-SFT from various financial data sources. These instructions encompass the following main categories: 





%


    
\begin{itemize}
    \item \textbf{Financial Consulting Instructions} constructed from financial Q\&A datasets and online financial forums; 
    \item \textbf{Financial Task Instructions} derived from existing and self-constructed NLP datasets; 
    \item \textbf{Financial Computing Instructions} based on a variety of financial statistical, computational and modeling problems;    
    \item \textbf{Retrieval-enhanced Instructions} built from financial texts with generated questions, retrieved references, and generated answers; 
\end{itemize}

Considering financial consultation, financial documents processing, financial computation problems resolving and financial knowledge retrieval are four entangled abilities, we utilize a Multiple Experts Fine-tuning Framework (MEFF) to build the intelligent financial system, named DISC-FinLLM, based on the constructed DISC-FIN-SFT instruction dataset. In specific, we train four individual Low-rank adaptation (LoRA)~\cite{hu2021lora} modules of our model on four parts of dataset respectively, which are designed to adopt multiple financial scenarios: financial multi-round dialogues, financial NLP tasks, financial calculation, and retrieval question answering. Therefore, these modules in our system can provide different services to corresponding user groups such as financial professionals, developers, and students to meet their specific needs, as shown in Figure.~\ref{fig:MEFF}. In this version, we use Baichuan-13B~\cite{Baichuan13B}, a general domain LLM for Chinese language, as the backbone.

In order to evaluate the effectiveness DISC-FinLLM, we utilize multiple evaluation benchmarks and experimental results show that DISC-FinLLM outperforms significantly better than the base foundation model in all downstream tasks. Further analysis demonstrates the advantage of our MEFF framework.

\begin{figure*}
      \centering
      \includegraphics[width=2.0\columnwidth]{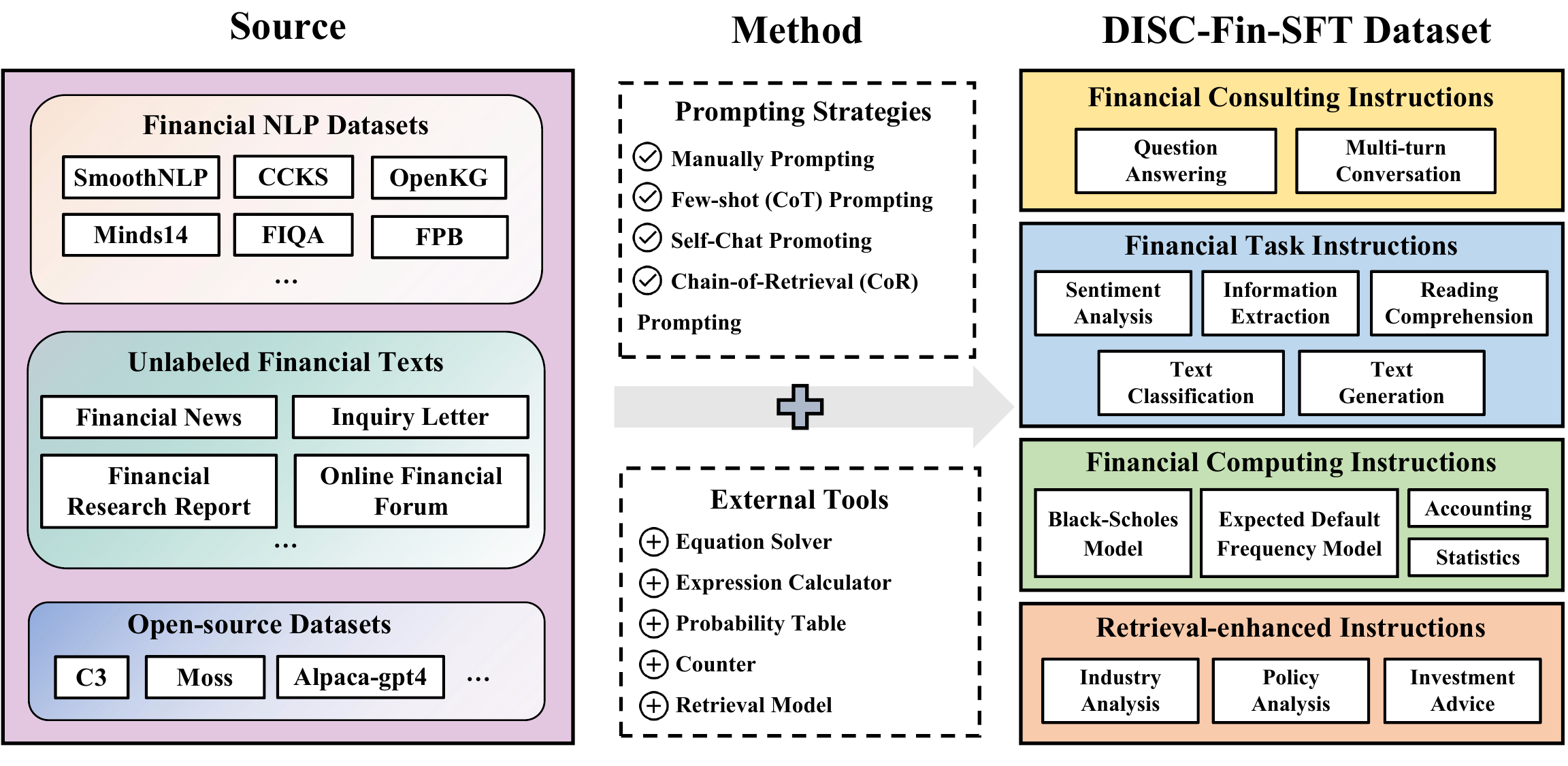}
      \caption{Construction of DISC-Fin-SFT Dataset.}
      \label{fig:dataset}
\end{figure*}


\section{Related Work}

\subsection{Financial NLP Models and Limitations}

Traditional financial NLP models have made progress in various financial scenarios such as named entity recognition~\cite{nakayama2017exploratory}, news sentiment analysis~\cite{souma2019enhanced,araci2019finbert}, event extraction~\cite{yang2018dcfee,zheng2019doc2edag}, report generation~\cite{chapman2022towards}, and text summarization~\cite{la2020end}. 

The application of NLP models to the financial sector presents a unique set of challenges. First, the intricate landscape of finance, which replete with complex terminology and rules, causes a shortfall in specialized knowledge~\cite{mik2017smart}. Second, the dearth of annotated data, coupled with costly annotation processes, impedes progress. Third, current NLP models exhibit limited inferential capacity, struggling with tasks like risk assessment and investment decision-making~\cite{liu2023fingpt}. Fourth, the rapid evolution of financial data demands real-time responsiveness, which these models often lack. Moreover, handling numerical computations within financial texts, abundant with numbers and symbols, presents another challenge. Finally, many NLP models show poor adaptability, being designed for single-task performance and lacking cross-task generalization~\cite{mishra2021cross}. These challenges underscore the need for future research to develop more robust and adaptable NLP models for the ever-evolving financial sector.

\subsection{Large Language Models for Finance}

The proposal of LLM-based dialogue systems like ChatGPT~\cite{openai2023chatgpt}, GPT-4~\cite{openai2023gpt4}, Alpaca~\cite{alpaca} have subverted previous dialogue systems~\cite{zhang2019dialogpt,chen2022contextual,chen2022dialogved}. These systems are famous for their zero-shot generalization ability~\cite{zhao2023survey}. One of the key technologies is instruction-tuning~\cite{wei2021finetuned}. Fine-tuning pre-trained LLM through diverse instruction data to obtain the desired behavior pattern has become a common way to domainize LLM~\cite{bao2023disc,yue2023disc}.

In financial field, the first notable example of domain-adapted LLM is BloombergGPT~\cite{wu2023bloomberggpt} based on BLOOM-176B~\cite{scao2022bloom}. It incorporates financial corpus during continue pre-training and exhibits promise in tasks such as financial forecasting and risk assessment. XuanYuan 2.0~\cite{zhang2023xuanyuan}, on the other hand, is the first Chinese financial LLM with hundreds of billions of parameters. It undergoes targeted pre-training and fine-tuning also using BLOOM-176B~\cite{scao2022bloom} for both the Chinese general field and the financial domain, showcasing exceptional performance in tasks like analysis and news comprehension. Nevertheless, the closed-source nature of the datasets used by BloombergGPT~\cite{wu2023bloomberggpt} and XuanYuan 2.0~\cite{zhang2023xuanyuan} poses challenges in developing financial LLMs. Additionally, their enormous parameter sizes result in high training costs. To address these limitations, FinGPT~\cite{liu2023fingpt} takes a data-centric approach, employing low-rank adaptation techniques while providing resources to researchers and practitioners for developing their own financial LLMs. However, FinGPT still necessitates fine-tuning on downstream tasks and is essentially similar to previous dedicated financial NLP models. PIXIU~\cite{xie2023pixiu} is a recent financial LLM closest to our work, while the training data of PIXIU only includes financial NLP datasets with human written prompts, which limits its wider application.

\section{DISC-Fin-SFT Datasets}




\begin{table}
\centering
\resizebox{\columnwidth}{!}{%
\begin{tabular}{rrrr} \toprule
\scshape Dataset    & \scshape \#Samples & \begin{tabular}[r]{@{}r@{}}\scshape Input \\ \scshape Length\end{tabular} & \begin{tabular}[r]{@{}r@{}}\scshape Output \\ \scshape Length\end{tabular} \\ \midrule
Consulting & 63k        & 26                                                      & 369                                                      \\
Task & 110k        & 676                                                     & 35                                                        \\
Computing & 57k        & 73                                                        & 190                                                        \\
Retrieval & 20k        & 1031                                                        & 521 \\ 
\midrule
Total & 246k        & 351                                                       & 198
\\ \bottomrule
\end{tabular}}
\caption{Data statistics of the DISC-Fin-SFT dataset. The input and output lengths are the average number of words after performing whitespace tokenization.}
\label{tab:statistics}
\end{table}



We construct DISC-Fin-SFT with approximately 250k examples derived from various sources, which mainly consist of 4 parts: financial consulting instructions, financial task instructions, financial computing  instructions and retrieval-enhanced instructions, as shown in Figure. \ref{fig:dataset}. Table. ~\ref{tab:statistics} provides details of the dataset.

\subsection{Financial Consulting Instructs} 




To construct instructions for financial consulting, we start with FiQA~\cite{maia201818} dataset, which is the only existing financial question answering (QA) dataset we are able to find. However, FiQA dataset is in English and the quality of the answers is not high enough. To leverage this dataset, we translate all questions in FiQA to Chinese and regenerate the corresponding answers using ChatGPT~\cite{openai2023chatgpt}. 

To enhance the LLM's understanding of financial terms, we collect over 200 finance-specific terms (e.g., Leveraged Buyout/LBO) from online sources and employ ChatGPT to generate corresponding QA pairs for these terms. 

In addition, we crawl posts from economic forum and financial investment forum on JingGuan\footnote{\url{https://bbs.pinggu.org/}}, an active Chinese financial forum. Self-chat prompting~\cite{xu2023baize} method is then utilized to guide ChatGPT to generate multi-turn QA centered around forum posting topic. 

The above prompts for answer generation are all carefully designed to ensure that the responses are consistent with Chinese national conditions, stance, attitude, language style, and other relevant aspects. For specific prompt templates, please refer to Appendix \ref{sec:prompts}.

\subsection{Financial Task Instructions}

In the financial field, there is a wide range of text-based financial tasks. We build financial task instructions mainly from two types of sources: existing financial NLP datasets and unlabeled financial texts. 


\paragraph{Financial NLP Datasets} ~ {We leverage existing financial NLP datasets with human-written prompts to build task instructions, following FLAN~\cite{wei2021finetuned}. We collect over 10 publicly available Chinese financial NLP datasets. These datasets can be categorized by task types as follows: 1) \textbf{Sentiment Analysis}, including FPB~\cite{malo2014good}, FiQA-SA~\cite{maia201818} and FNSC\footnote{\url{https://github.com/wwwxmu/Dataset-of-financial-news-sentiment-classification}}; 2) \textbf{Information Extraction}, including FR-NER~\cite{jia2020entity}, OpenKG~\cite{ren2022iree}, CCKS-NEC-2022~\cite{ccksnec2022} and SmoothNLP\footnote{\url{https://github.com/smoothnlp/FinancialDatasets}}; 3) \textbf{Text Classification}, including Minds14~\cite{gerz2021multilingual} and CCKS-2022~\cite{ccks2022event}; 4) \textbf{Text Generation}, including SmoothNLP and Finance-alpaca-KG\footnote{\url{https://huggingface.co/datasets/gbharti/finance-alpaca}}. In addition, we include C3~\cite{sun2020investigating}, a general domain Chinese multiple-choice QA dataset, to adapt to the subsequent evaluation mechanism. We summarize the details of each dataset in Appendix~\ref{sec:nlp_task}. We hand-code more than 20 prompt templates for each dataset and manually write prompts for both zero-shot and few-shot scenarios for all non-generative tasks. One of the prompt templates can be seen in Figure. \ref{3}. This ensures the LLM to retain the in-context learning ability while enhancing the zero-shot ability. 


%



%


}

\paragraph{Unlabeled Financial Texts} ~ {To enhance the understanding of financial texts in real scenarios, we construct a reading comprehension dataset using unlabeled financial texts. We first collect a total of 87k passages consisting of 69k financial news and 18k industry research report summaries from East Money~\footnote{\url{https://www.eastmoney.com/}}, a reputable internet financial media, spanning from January 1 to August 16, 2023. These financial passages encompass a range of topics, including company disclosures, coverage of events, macroeconomic analysis, and industry research, etc. Since the passages are generally long, we further use the sentence segmentation algorithm to divide each passage into multiple  paragraphs, and finally generate a total of 1.8 million paragraphs as the basic units. Furthermore, for a given paragraph, we utilize ChatGPT~\cite{openai2023chatgpt} to generate QA pairs by the prompt in Figure. \ref{fig:question} and \ref{4} to obtain (paragraph, question, answer) triplets, which are then utilized to build instructions using diverse reading comprehension task templates.

}

\begin{figure*}
      \centering
      \includegraphics[width=0.95\textwidth]{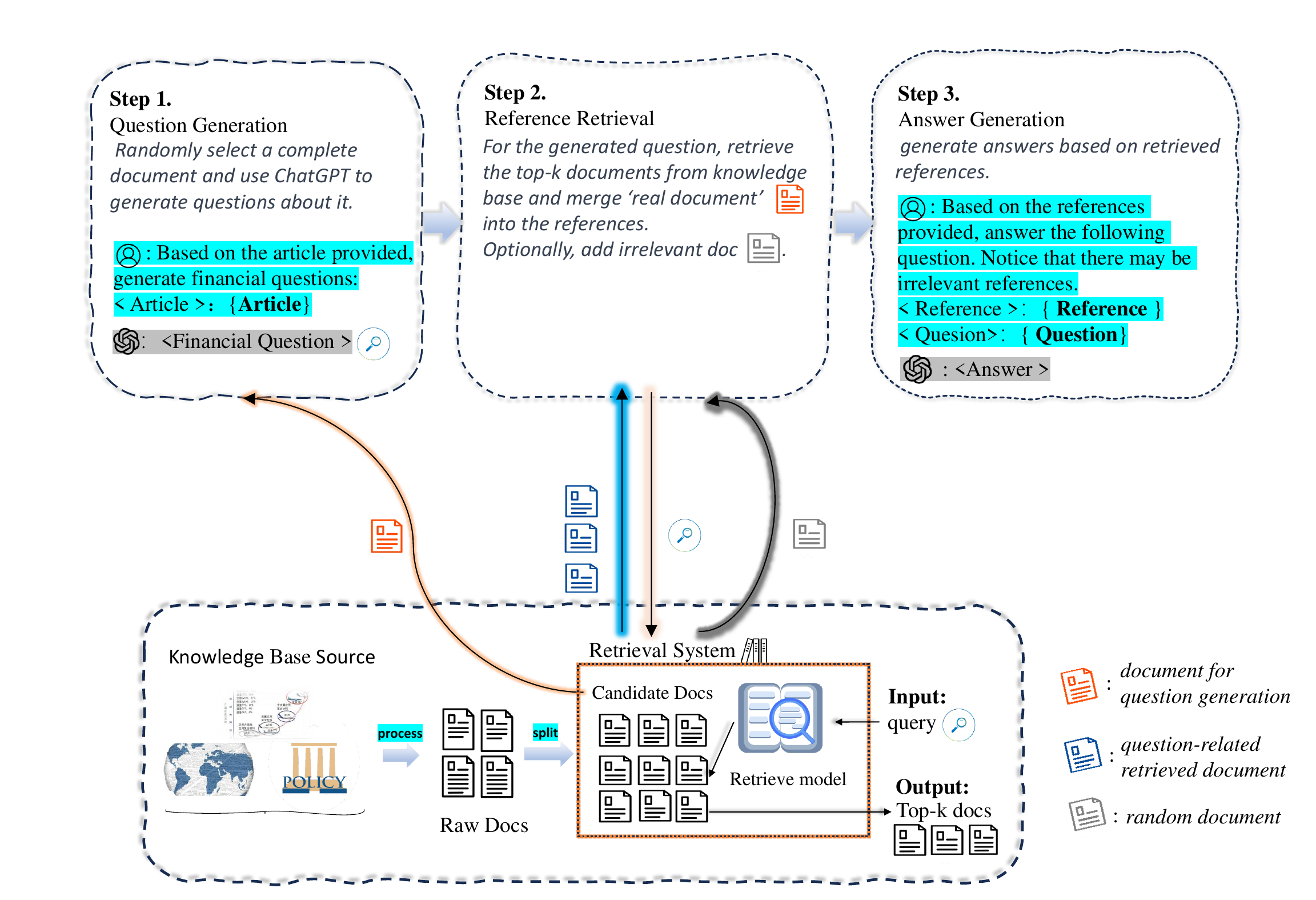}
      \caption{Process diagram for constructing retrieval instructions.}
      \label{6}
\end{figure*}

\subsection{Financial Computing Instructions}

Financial texts, especially financial reports, are filled with a large amount of numerical information. In the process of digital analysis, we inevitably need to perform some calculations, such as growth rate, expected earnings, etc. In order to support this function in large language models, we construct calculation plugin instruction data. 

In the financial field, the four tools shown in Table. ~\ref{2} can meet most computing tasks, and their calling commands, inputs and outputs are different. For example, the command of the calculator is [Calculator(expression)→result].

\begin{table}
    \centering
    \renewcommand\arraystretch{1.15}
    \resizebox{\columnwidth}{!}{
    \begin{tabular}{ll} \toprule
    
        \scshape Tool             &  \scshape Detail  \\ \midrule
    Expression calculator     & Input: expression\\
&Output: result\\\midrule
    Equation solver &Input: equation system\\
&Output: solution \\ \midrule
    Counter       & Input: array of samples\\
&Output: sample size              \\ \midrule
    Probability table  & Input: number\\

 &Output: cumulative standard\\
&normal distribution function \\
& value at this number\\
    \bottomrule
    \end{tabular}
    }
\caption{Definition of calculation tools}
\label{2}
\end{table}

%
%
    
    
    
    
    
    
    

We first constructed a seed task pool, which consists of three parts: handwritten financial calculation questions based on the financial exams, arithmetic questions with contexts of financial research reports, and general mathematical questions in Belle School Math~\cite{ji2023exploring}. In particular, the answers to these questions are inserted into calling commands of the above four tools that represent when the tool should be called and how the tool should be used, according to the method of Toolformer~\cite{Toolformer}. Subsequently, for increasing the data amount and diversity, we utilize ChatGPT~\cite{openai2023chatgpt}  to generate more than 50,000 new calculation question-answer pairs by self-instruction~\cite{selfinstruct} and few-shot Chain-of-Thought (CoT) prompting based on the seed tasks, of which the answers also come with plugin commands. Figure. \ref{5} illustrates the prompt template for ChatGPT when generating financial computing instructions.

Through training on this dataset, the model can learn to comprehensively apply the above four tools at the appropriate time to assist it in completing computing tasks.


\subsection{Retrieval-enhanced Instructions} 


In order to further improve retrieval-enhanced generation capabilities, particularly in the context of finance, encompassing aspects such as professionalism, adept utilization of reference materials, critical reasoning, and creative expression, we have employed a 3-step methodology for constructing retrieval-enhanced instructional data. This instructional data comprises financial domain inquiries, pertinent reference documents, and their respective responses. As shown in Figure. \ref{6}, the three main steps of the methodology are as follows:

\begin{enumerate}
    \item[1)] Question Generation. Formulate financial analysis questions derived from financial materials such as news articles and research reports.
    \item[2)] Reference Retrieval. Retrieve documents from our knowledge base that are germane to the questions with a predefined threshold.
    \item[3)] Answer Generation. Merge the generated questions with the retrieved reference materials to produce suggestions.
\end{enumerate}

Both questions and answers are generated by ChatGPT through Chain-of-Retrieval (CoR) prompting. The reference documents mentioned in step 2 are sourced from our proprietary financial knowledge base, comprising 18,000 abstracts of research reports and 69,000 financial news articles spanning from January 1, 2023, to August 16, 2023. To bolster the model's capacity to discern and filter out irrelevant text, we randomly introduce non-relevant documents alongside retrieved materials, as discussed in~\cite{lawyer-llama-report}. To mitigate the shortcomings of external retrieval models, we randomly incorporate the documents used for question generation when they are not retrieved by the system. 

Ultimately, we develop a corpus of 20k Retrieval-enhanced Instructions, covering prevalent analysis categories within the realm of finance, namely \textit{industry analysis} (53\%), \textit{policy analysis} (13\%), \textit{investment guidance} (8\%) and other financial scenarios such as \textit{corporate strategic planning, Technical Analysis}  (26\%).Our instructional data plays a crucial role in guiding the language model to effectively harness reference documents, encompassing aspects like enhancing its comprehension, summarization capabilities, and the identification of irrelevant documents. It is worth noting that when we actually use CoR prompting, the Prompt used is more complicated. Please refer to Figures. ~\ref{fig:news}, \ref{fig:reports}, and \ref{fig:answer} for details.

\section{Multiple Experts Fine-tuning Framework}

\begin{figure}
      \centering
      \includegraphics[width=0.95\columnwidth]{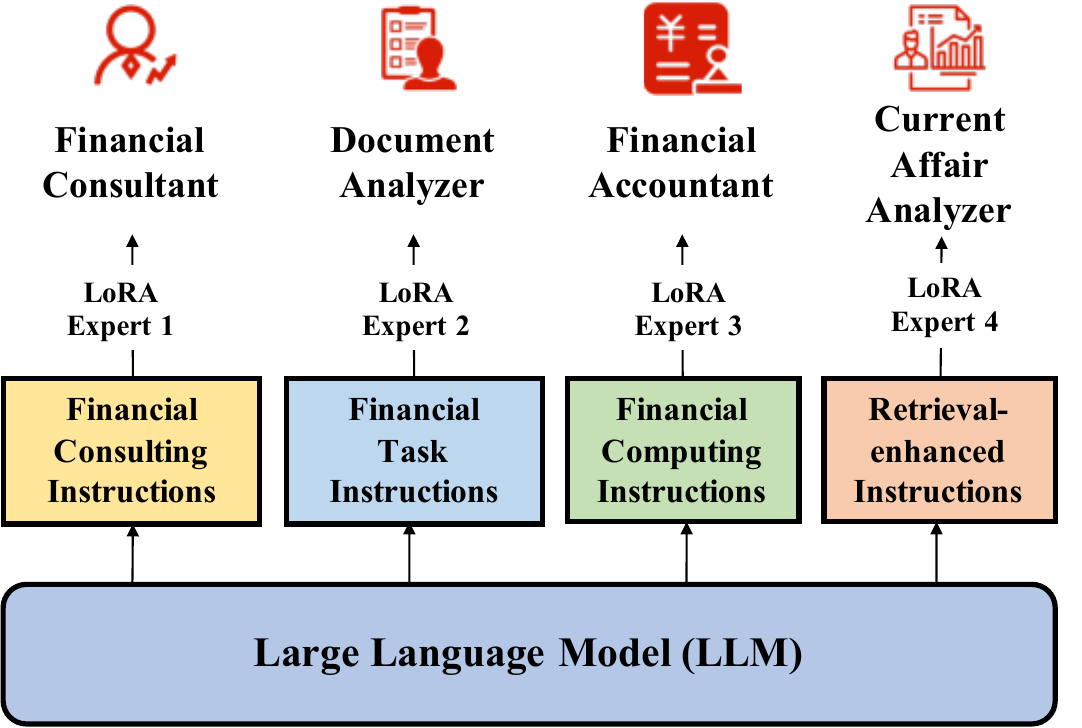}
      \caption{Multiple experts fine-tuning framework of DISC-FinLLM.}
      \label{fig:moe}
\end{figure}

In this section, we present the methodology employed in the construction of financial large language models, as shown in Figure.~\ref{fig:moe}.

\subsection{Overall Architecture}

To ensure the specialized and task-oriented functionality in our system, we propose a Multiple Experts Fine-tuning strategy. This strategy is tailored to address distinct functionalities within the financial domain. We train individual components of our model on specialized datasets, allowing them to operate autonomously without interfering with one another. To accomplish this, we leverage the Low-rank adaptation (LoRA)~\cite{hu2021lora} mechanism for efficient parameter fine-tuning. 

Specifically, we train 4 LoRA modules by fine-tuning base model on the four parts of instructions. During the deployment, switching between different features simply involves replacing the LoRA parameters loaded on the current base model. This enables us to activate/deactivate different functionalities of the model based on the task requirements without retraining the entire model. Moreover, this approach enhances the model's efficiency and leads to improved performance. 

\subsection{Multiple Experts Setup}

Given the multifaceted nature of the financial domain, we have primarily curated four distinct categories of financial datasets, including financial consulting instructions, financial task instructions, financial computing instructions, and retrieval-enhanced instructions. Consequently, we train LoRA models tailored to each of these four directions. 



\paragraph{Financial Consultation} This first LoRA model aims to address conversational challenges, with a particular focus on question-answering tasks within the financial domain. The strength of this model in responding to queries, especially those related to finance, can be attributed to the abundance of data encompassing financial questions and consultations tailored to the Chinese financial context. 
 
\paragraph{Financial Document Processing} The second LoRA model is primarily engineered to tackle various tasks within the realm of financial natural language processing. These tasks include, but are not limited to, financial information extraction and sentiment analysis in financial texts. 

 
\paragraph{Financial Computing} The third LoRA training is to acquire calculation plug-in. DISC-FinLLM supports four tools, namely expression calculator, equation solver, counter, and probability table. These tools support our model to complete major calculation tasks in the financial field, such as financial mathematical modeling, statistical analysis, etc. When the model needs to use tools, it can generate tool call commands, then interrupt decoding, and add the tool call results to the generated text. In this way, DISC-FinLLM can accurately solve arithmetic problems in finance based on the calculation results provided by the tools. 
 
\paragraph{Financial Knowledge Retrieval} The fourth LoRA training process aims to inject retrieval plug-in. DISC-FinLLM improves retrieval-enhanced generation capabilities mainly in three financial domains: news, report and policy. When asked about common financial topics such as current affairs, industry trends or financial policies, our model can retrieve relevant documents and analyzing them like a financial expert, and ultimately providing suggestions.

\begin{table*}[]
\centering
\resizebox{0.9\textwidth}{!}{%
\begin{tabular}{rccccccc} \toprule
\begin{tabular}[c]{@{}c@{}}\scshape Eval on →\\\scshape Model   
 ↓\end{tabular} &
  \begin{tabular}[c]{@{}c@{}}\scshape FinFE\\ \scshape (Acc)\end{tabular} &
  \begin{tabular}[c]{@{}c@{}}\scshape FinQA\\ \scshape (F1)\end{tabular} &
  \begin{tabular}[c]{@{}c@{}}\scshape FinCQA\\ \scshape (F1)\end{tabular} &
  \begin{tabular}[c]{@{}c@{}}\scshape FinNA\\ \scshape (ROUGE)\end{tabular} &
  \begin{tabular}[c]{@{}c@{}}\scshape \scshape FinRE\\ \scshape (F1)\end{tabular} &
  \begin{tabular}[c]{@{}c@{}}\scshape FinESE\\ \scshape (F1)\end{tabular} &
  \begin{tabular}[c]{@{}c@{}}\scshape Avg\end{tabular} \\ \midrule
Baichuan-13B-Chat  &   64.8                                                                      &        38.1                                 &             33.6                                                  &         \textbf{31.0}                                                          &        9.1                                                                   &           18.6                      & 31.0                                          \\
(LoRA)                                                       &             \textbf{69.3}                                                               &               \textbf{42.4}                                                        &          \textbf{42.0}                                                              &      30.9                                                                      &    \textbf{10.1}                                                                       &                   \textbf{45.3}      &   \textbf{40.0}                                                               \\  \midrule
ChatGLM                                                                &         56.7                                                                        &          31.8                                                                &        35.1                                                                      &       32.5                                                                          &           \textbf{13.0}                                                                        &                   \textbf{48.7}               & 36.3                                          \\
(LoRA)                                                              &       \textbf{60.7}                                                                         &         \textbf{41.4}                                                                 &             \textbf{36.4}                                                              &       \textbf{34.7}                                                   &   10.7                                                                       &            46.2     &  \textbf{38.4}     \\  \midrule
ChatGLM2                                                                &                   61.3                                                     &            28.8                                                           &         35.9                                                             &        28.9                                                                    &        11.7                                                                     &                                    \textbf{42.1}    & 34.8                                                             \\
(LoRA)                                                              &                \textbf{65.3}                                                       &                  \textbf{37.6}                                                  &            \textbf{36.4}                                                     &            \textbf{33.4}                                                                  &     \textbf{11.8}                                                               &          39.5       &    \textbf{37.3}                                                             \\  \bottomrule
\end{tabular}}
\caption{Experimental results on the BBT-FIN benchmark.}
\label{tab:bbt}
\end{table*}

\begin{table*}
\centering
\resizebox{0.9\textwidth}{!}{%
\begin{tabular}{lrrrrr} \toprule
\scshape Model              & \scshape Finance & \scshape Economy & \scshape Accounting & \scshape Certificate & \scshape Avg  \\ \midrule
GPT-4               & 71.0                          & 74.5                        & 59.3                           & 70.4                            & 68.6                         \\
ChatGPT           & 59.3                        & 61.6                        & 45.2                           & 55.1                            & 55.0                           \\ \midrule
Baichuan-13B-Base   & 52.6                        & 50.2                        & 43.4                           & \textbf{53.5}                            & 50.1                         \\
Baichuan-13B-Chat  & 51.6                        & \textbf{51.1}                        & 41.7                           & 52.8                            & 49.4                         \\
ChatGLM2-6B     & 46.5                        & 46.4                        & 44.5                           & 51.5                            & 47.4                         \\
InternLM-7B        & 49.0                          & 49.2                        & 40.5                           & 49.4                            & 47.1                         \\
InternLM-Chat-7B     & 48.4                        & 49.1                        & 40.8                           & 49.5                            & 47.0                           \\
LLaMA-2-Chat-70B    & 47.1                        & 46.7                        & 41.5                           & 45.7                            & 45.2                         \\ 
FinGPT-v3-6B   & 50.5                       & 42.5                        & 50.8                          & 52.1                           & 49.6                         \\ \midrule
DISC-FinLLM (Consulting)  & 54.4                       & 45.4                        & \textbf{52.8}                          & 51.8                           & \textbf{51.6}                        \\ 
DISC-FinLLM (Task)              &    \textbf{57.4}                        &  48.8                      &    49.5                     &    49.7                         & 51.5              \\
DISC-FinLLM (Retrieval)    & 56.1                       & 44.0                        & 49.5                         & 50.6                           & 50.6                         \\ 
DISC-FinLLM (Computing)  & 54.8                       & 50.2                        & 46.9                       & 50.6                           & 50.9                        \\ 
 \midrule 
\textbf{Ablation Study}  \\   \midrule
DISC-FinLLM (full)    & 53.8                       &  47.9                       &  42.0                       &  49.1                           & 48.7                         \\  \bottomrule          
\end{tabular}}
\caption{Experimental results on the FIN-Eval benchmark.}
\label{tab:fineval}
\end{table*}



\begin{table*}
\centering
\begin{tabular}{lcc} \toprule
               \scshape Model    & \scshape Formula &  \scshape Formula \& Result  \\ \midrule
GPT-3.5-turbo     & 0.28      & 0.26               \\
Baichuan-13B-Chat & 0.20      & 0.12                \\ \midrule
DISC-FinLLM (Computing)       & 0.35      & 0.35              \\ \bottomrule
\end{tabular}
\caption{Evaluation results of calculation plugin.}
\label{tab:computing}
\end{table*}

\begin{table*}
\centering
\begin{tabular}{lcccc} \toprule
                 \scshape Model & \scshape Accuracy & \scshape Usefulness & \scshape Linguistic & \scshape Reflectiveness  \\ \midrule
Baichuan-13B-Chat & 4.08     & 4.15       & 4.21               & 3.88            \\
DISC-FinLLM (Retrieval)      & 4.13     & 4.29       & 4.33               & 3.95   \\ \bottomrule        
\end{tabular}
\caption{Evaluation results of retrieval plugin.}
\label{tab:retrieve}
\end{table*}


\section{Experiments}

\subsection{Evaluation Setup}


We establish a comprehensive evaluation framework to assess our financial large language model from various perspectives. It comprises four components, namely: financial NLP tasks, human tests, data analysis, and current affairs analysis. 

\paragraph{Financial NLP Tasks} ~ {To assess model's NLP ability, we utilize the FinCUGE evaluation benchmark~\cite{2023arXiv230209432L}. We evaluate six of these tasks, which include sentiment analysis, relation extraction, summarization, text classification, event extraction, and other tasks. These six tasks correspond to six datasets, namely FinFE, FinQA, FinCQA, FinNA, FinRE and FinESE. You can find detailed information about these datasets in the Appendix ~\ref{sec:bbt6}. To create a few-shot evaluation setting, we transform the test set by providing prompts. We measure the performance using accuracy, F1 score, and rouge score. }

\paragraph{Human Tests} ~ {To evaluate our model's performance on human-generated financial questions, we employ the FinEval benchmark~\cite{zhang2023fineval}. It is a collection of high-quality multiple-choice questions covering finance, economy, accounting, and certificate. We employ a few-shot approach to measure the performance of various models using the accuracy metric.}

\paragraph{Data Analysis} ~ {For evaluating our model's capabilities in computational tasks, we manually create a dataset consisting of over 100 financial calculation problems, which are adapted from material analysis computational questions in Chinese Administrative Aptitude Test. The dataset is created entirely through manual efforts to ensure the quality. We assess the model's performance by the metric of accuracy in terms of formula construction and results calculation respectively.}

\paragraph{Current Affairs Analysis} ~ {To evaluate our model's performance in retrieval-based tasks, we design a dataset of financial questions to require the use of up-to-date information for accurate answers. Reference documents are collected manually based on search engine. This dataset enables us to evaluate the model's ability to retrieve relevant and precise information when presented with specific financial queries. We use GPT-3.5 to evaluate the generated results according to four metrics, namely, accuracy, usefulness, linguistic quality and reflectiveness.
\begin{itemize}
    \item Accuracy: The provided recommendations or analysis are accurate, with no factual errors, and conclusions are not arbitrary
    \item Usefulness: It can provide clear and practical analysis and opinions on issues in the financial domain, in conjunction with the reference text.
    \item Linguistic Quality: It can correctly understand the questions and generate concise, professional answers within the financial domain.
    \item Reflectiveness: It can analyze and reflect on reference documents, summarize and derive conclusions rather than simply copying from the original text.
\end{itemize}
\noindent
}

\subsection{Main Results}

In this section, we present evaluation results of our model on various tasks in the financial domain.

\paragraph{Financial NLP Tasks}  Table. \ref{tab:bbt} shows results on the set of financial NLP tasks. We conduct LoRA training on financial task instructions separately for Baichuan-13B-Chat, ChatGLM, and ChatGLM2 models. We compare the evaluation results between the models before and after training. The experimental results show that our average performance on the six unseen tasks is 2 to 9 points higher than that of the untrained base model. Furthermore, it is important to note that certain NLP tasks included in the evaluation datasets were not covered by our own dataset. In such instances, these findings underscore the effectiveness of the specific task instruction data we have constructed in enhancing the model's generalization performance in financial domain.


\paragraph{Human Tests} Table. \ref{tab:fineval} presents the evaluation results of our model on human-generated financial questions. We conduct separate tests on the four LoRA-trained models and the model fine-tuned with complete data. The compared models include ChatGPT, GPT-4, Baichuan-13B, ChatGLM2, FinGPT-v3 and others. We present evaluation results on four LoRA-trained models (consulting, task, retrieval and computing). Our model achieves the best performance among all evaluated models in terms of average results, except for ChatGPT and GPT-4. our evaluation of FinGPT's results reveals that, in comparison to existing financial large language models, our model exhibits superior performance. We also use all the data to perform full-parameter training on the Baichuan-13B-Chat base model and conduct ablation study. The notable decrease in the evaluation results obtained after fine-tuning on the base model emphasizes the necessity of our task-specific LoRA fine-tuning approach for each task.

\paragraph{Data Analysis} Table. \ref{tab:computing} showcases the experiment results on financial computing tasks. We compare the evaluation results of the model that undergo LoRA training using financial computing instructions with those of the untrained Baichuan-13B model. The addition of computational plugins to our model generates a notable performance boost compared to the baseline models, surpassing ChatGPT by 0.09 points. These results highlight the efficacy of our approach in addressing computational challenges within the financial domain.

\paragraph{Current Affairs Analysis} Table. \ref{tab:retrieve} demonstrates the experiment results on retrieval-based test sets. Based on the evaluation results given by GPT-3.5, our model shows significantly higher results in all four metrics, whether accuracy, usefulness, linguistic quality or reflectiveness. These results prove the effectiveness of retrieval-enhanced instructions.

\section{Conclusion}

In this paper, we propose a multiple expert fine-tuning framework for building a powerful Chinese intelligent system in financial domain. We fine-tune our model on task-specific instruction data using LoRA and incorporate system prompts to enhance its performance in financial NLP tasks, human test evaluation, computational tasks, and retrieval tasks. Our evaluation results demonstrate the effectiveness of our model across these domains. Our model's strong performance opens up possibilities for applications in financial customer support, investment analysis, and risk assessment.

\bibliography{anthology,custom}
\bibliographystyle{acl_natbib}

\appendix


\section{Prompt Engineering}
\label{sec:prompts}

In this section, we summarize the prompts to generate the proposed financial instruction-tuning dataset DISC-FinLLM-SFT. 

\begin{figure}
      \centering
      \includegraphics[width=1.0\columnwidth]{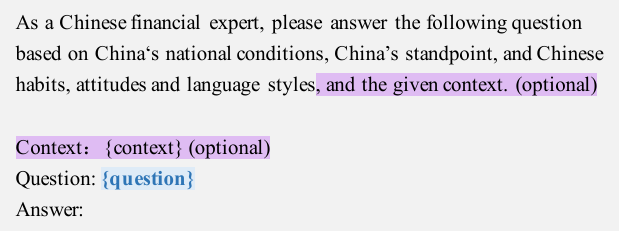}
      \caption{Zero-shot prompt template for generating answer to given financial question and optional context in financial consulting instructions.}
      \label{fig:question}
\end{figure}

There are two main functions of our templates, one is question generation~\cite{yao2021question,abdelghani2023gpt} and the other is answer generation~\cite{savelka2023large}. In Figure~\ref{fig:question}, the template is utilized to produce answers when provided with a question and optional context, without any demonstration examples (zero-shot). 

\begin{figure}
      \centering
      \includegraphics[width=1.0\columnwidth]{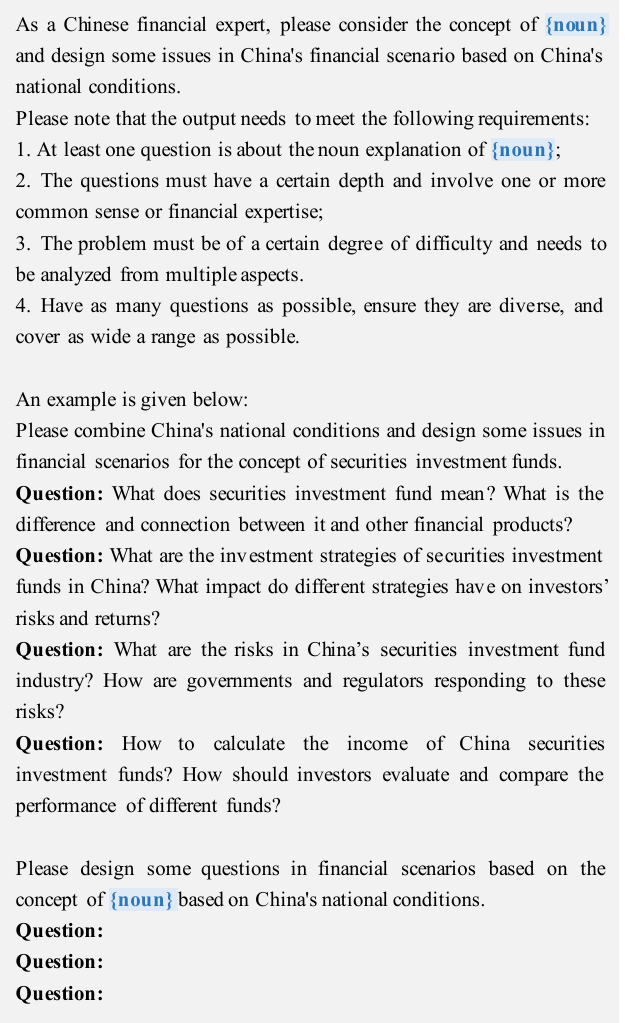}
      \caption{Few-shot prompt template for generating questions related to a financial terminology in financial consulting instructions.}
      \label{fig:term}
\end{figure}

\begin{figure}
      \centering
      \includegraphics[width=1.0\columnwidth]{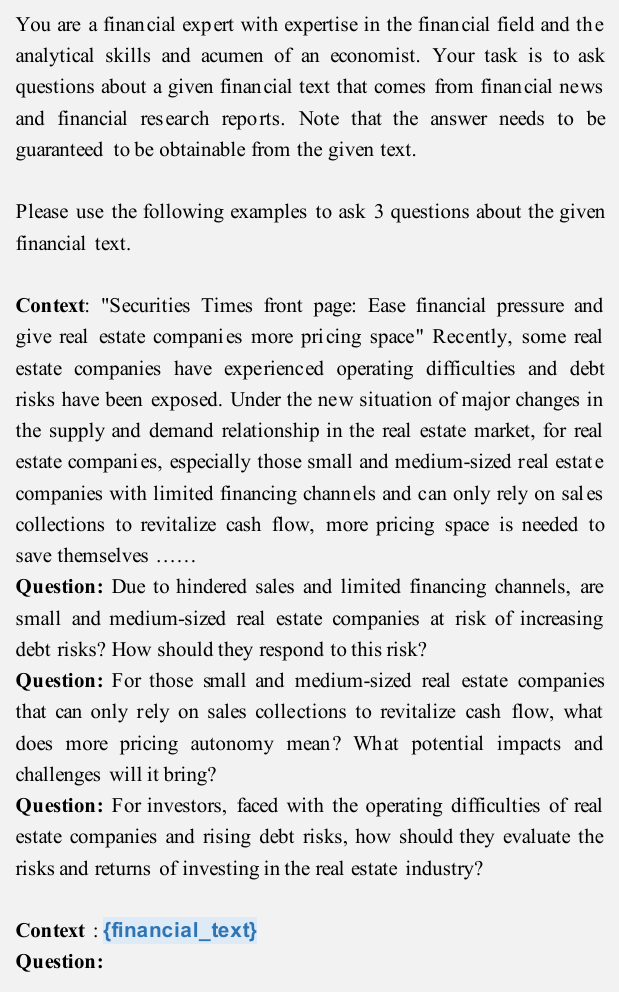}
      \caption{Few-shot prompt template for generating questions from unlabeled financial text in financial consulting instructions.}
      \label{4}
\end{figure}

The prompts in Figures~\ref{fig:term} and~\ref{4} are employed to generate questions from unlabeled financial text and given financial terminologies, respectively, in a few-shot manner. The process of constructing instructions for terminology Q\&A involves initially generating questions based on the given terms and subsequently generating answers. Similarly, our approach to constructing the reading comprehension dataset entails generating questions from unlabeled financial texts and then further generating answers.

\begin{figure}
      \centering
      \includegraphics[width=1.0\columnwidth]{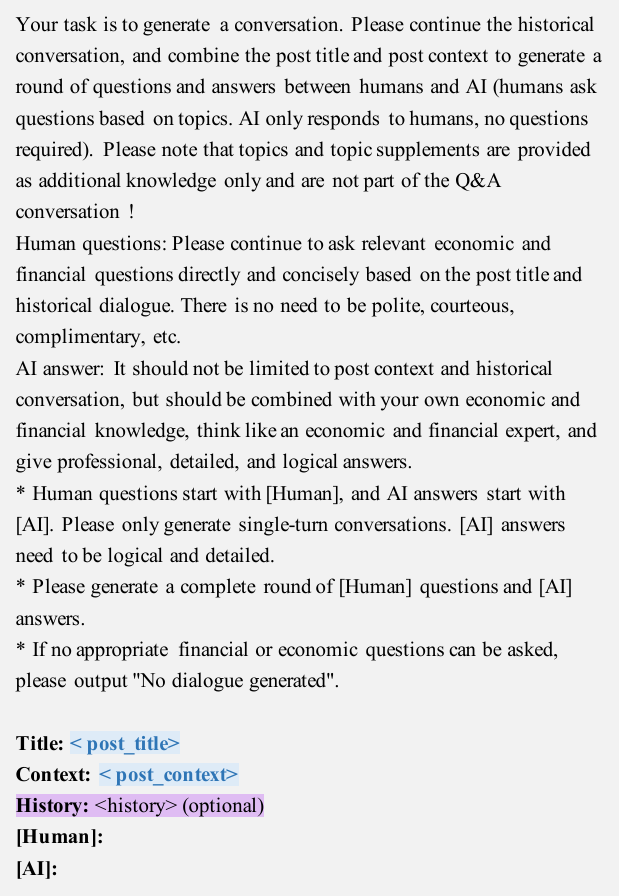}
      \caption{Zero-shot prompt template for generating multi-turn question answering instructions.}
      \label{fig:multi-turn}
\end{figure}

Regarding multi-turn question answering, we adopt a method similar to Baize~\cite{xu2023baize}. However, we observe that directly applying this method results in generated responses that are too short and lack detail. Therefore, we employ an iterative generation approach, generating one turn of human-AI dialogue at a time, as shown in Figure.~\ref{fig:multi-turn}. Human questions are required to be as direct and concise as possible, incorporating the given topic (from the post title) and the historical dialogue, without the need for politeness, compliments, or flattery. AI responses are expected to integrate contextual knowledge (from the post context) and the historical dialogue to provide professional, detailed, and logically consistent answers. 

\begin{figure}
      \centering
      \includegraphics[width=1.0\columnwidth]{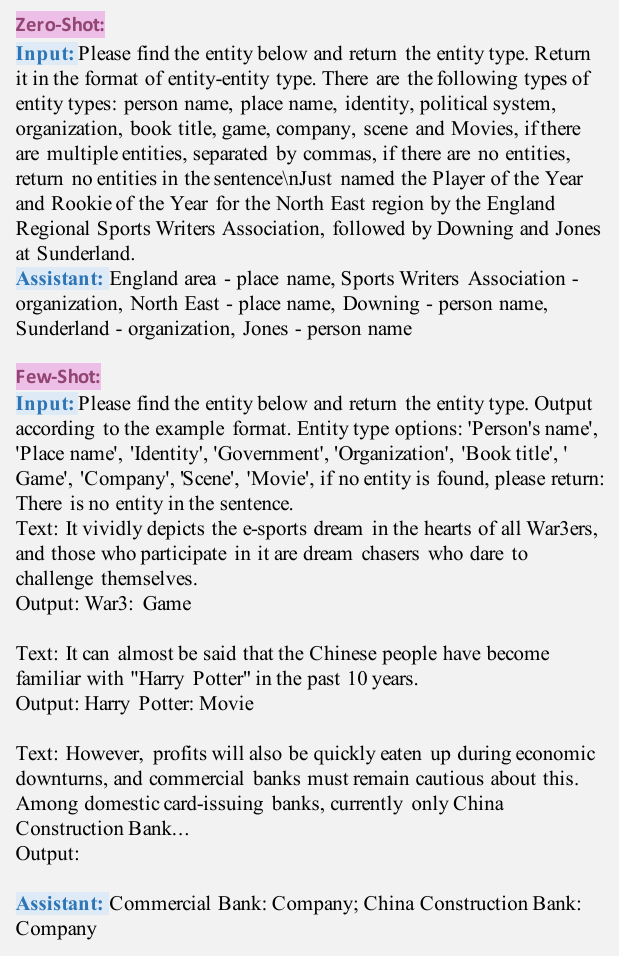}
      \caption{Zero-shot and few-shot prompt templates for constructing financial NLP task instructions.}
      \label{3}
\end{figure}

\begin{figure}
      \centering
      \includegraphics[width=1.0\columnwidth]{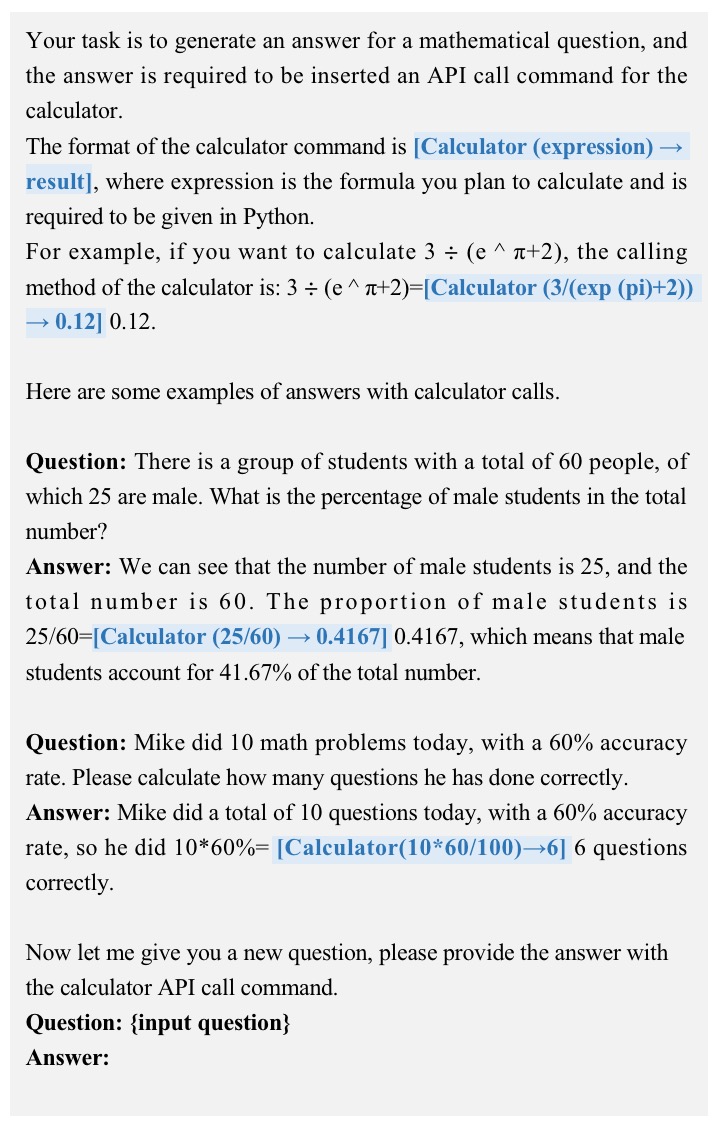}
      \caption{Prompt template for constructing financial computing instructions.}
      \label{5}
\end{figure}

\begin{figure}
      \centering
      \includegraphics[width=1.0\columnwidth]{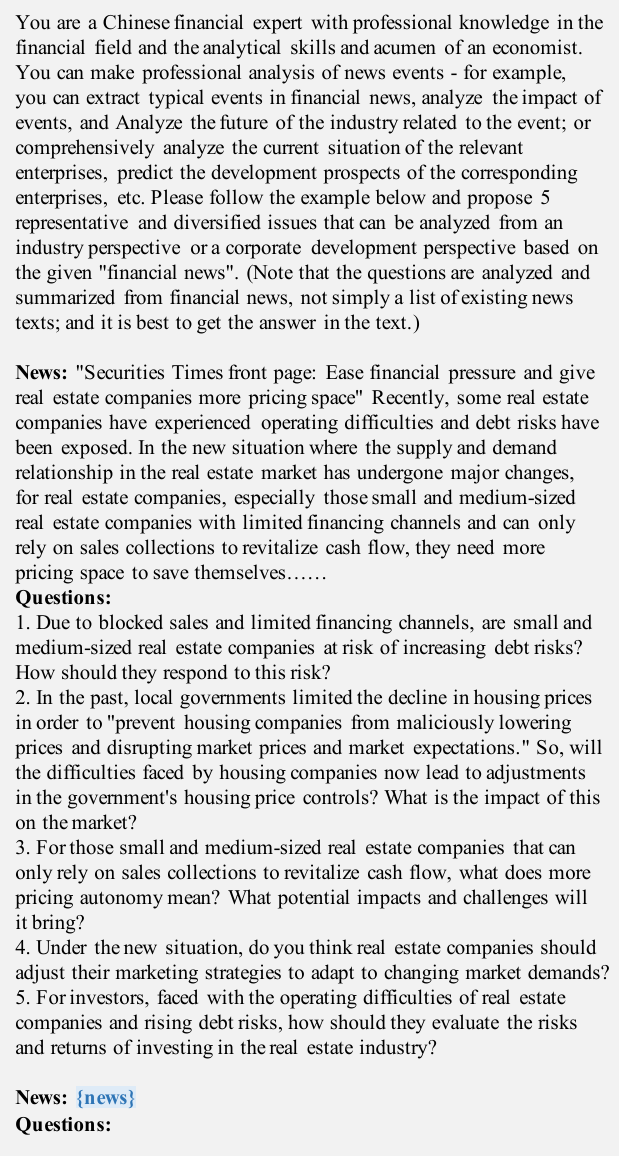}
      \caption{Few-shot prompt template for generating questions from financial news in retrieval-enhanced instructions.}
      \label{fig:news}
\end{figure}

\begin{figure}
      \centering
      \includegraphics[width=1.0\columnwidth]{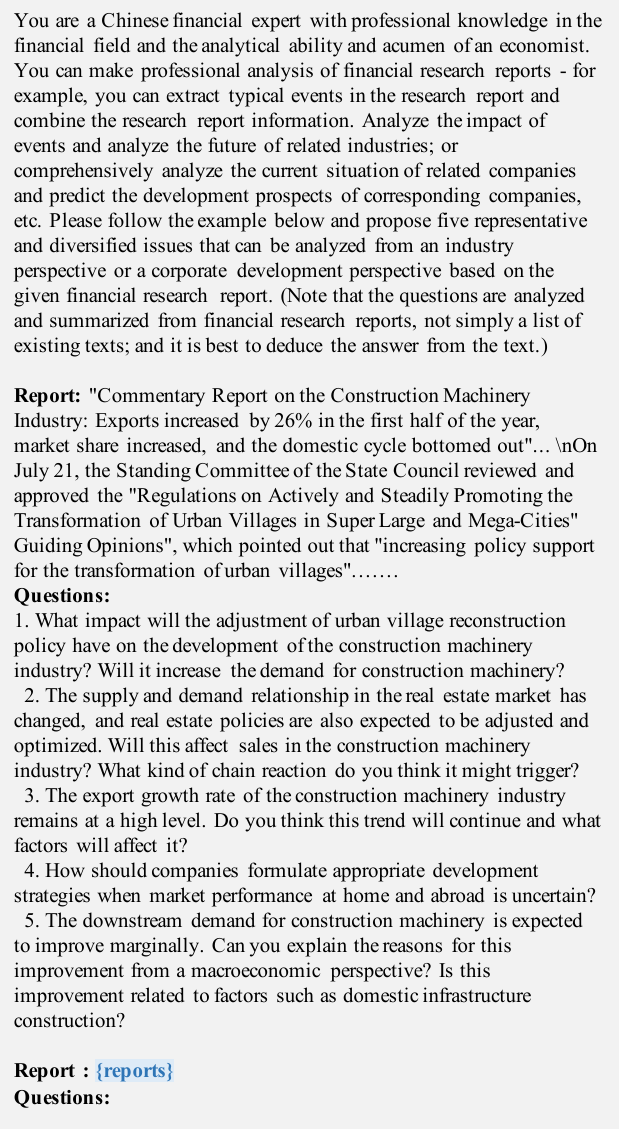}
      \caption{Few-shot prompt template for generating questions from financial research report in retrieval-enhanced instructions.}
      \label{fig:reports}
\end{figure}

\begin{figure}
      \centering
      \includegraphics[width=1.0\columnwidth]{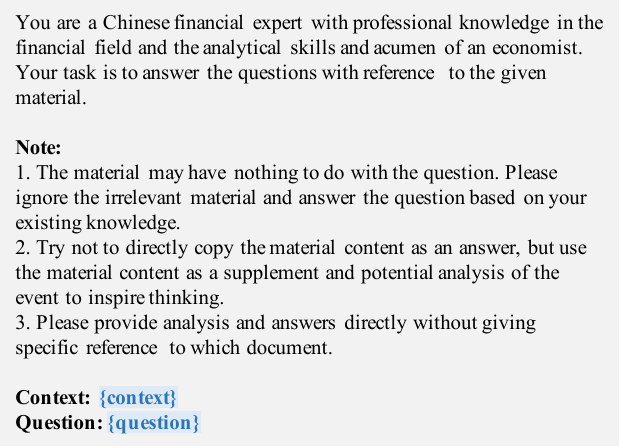}
      \caption{Zero-shot prompt template for generating answer to given financial question in retrieval-enhanced instructions.}
      \label{fig:answer}
\end{figure}

In Figure.~\ref{3}, For each financial NLP dataset for non-generative tasks, we construct both zero-shot and few-shot prompt templates. Figure.~\ref{5} shows a few-shot chain-of-thought (CoT) prompt for invoking computational tools. Figure .~\ref{fig:news} and .~\ref{fig:reports} show the few-shot prompt templates for generating questions from financial news and research reports respectively when building retrieval-enhanced instructions. Figure .~\ref{fig:answer} shows the prompt template for generating answer given the retrieved documents and question.



\section{Details of Financial NLP Task Instructions}
\label{sec:nlp_task}


In this section, we go into more detail about the financial NLP datasets used to construct our instructions. These datasets cover various tasks, including information extraction, sentiment analysis, text generation, and question answering, as shown in Figure .~\ref{tab:nlp_datasets}. 

\paragraph{Information Extraction} ~ {Information extraction is an important task of financial NLP, which is a key technology for constructing structured information from unstructured financial texts. We use the following dataset: 1) Financial Report~\cite{jia2020entity}, a dataset for named entity recognition (NER), which requires identifying 10 predefined entity categories from bank financial reporting; 2) SmoothNLP; 3) CCKS; 4) OpenKG; 5) minds14. The Financial Report dataset is designed for the task of selecting the most appropriate event type from a given set of event types based on provided financial texts. The SmoothNLP dataset includes subsets focused on investment event extraction, which has been structured into instructional data formats. These datasets primarily serve the purpose of information extraction. The CCKS dataset not only pertains to the analysis of event types within financial texts, but also involves causal relationship extraction. The OpenKG dataset involves the extraction of entities such as event types, time, subjects, and numerical values. Lastly, the minds14 dataset leverages speech data to accomplish the task of intent detection.}

\paragraph{Sentiment Analysis} ~ {Sentiment analysis helps to understand people's attitudes and sentiment towards financial products and services. We collect 3 datasets including FPB~\cite{malo2014good}, FiQA-SA~\cite{maia201818} and FNSC. Among them, FPB and FiQA-SA are English datasets, which we translated into Chinese using ChatGPT. FPB encompasses English sentences from financial and economic news articles, along with expert annotations providing sentiment labels categorized as positive, negative, or neutral. The goal of FiQA-SA dataset is to predict the sentiment of English financial and economic news articles and microblog posts. For this dataset, we have chosen two types of sentiment labels: discrete positive and negative categories, as well as continuous values within the range of (-1, 1). In this range, a label of 1 indicates the most positive sentiment, while -1 represents the most negative sentiment. This approach is tailored to facilitate enhanced sentiment analysis by models in the financial domain, enabling a finer-grained understanding of emotional nuances. Additionally, the FNSC dataset focuses on sentiment classification within financial news, featuring two labels: positive and negative. }

\paragraph{Text Classification} ~ {We select the CCKS-2022~\cite{ccks2022event} dataset for topic classification to identify the specific professional fields to which the text content belongs, and the Minds14~\cite{gerz2021multilingual} dataset for intent identification to identify the intent of bank users' spoken words.}

\paragraph{Text Generation} ~ {Our text generation task types include news headline generation and keyword generation. The datasets utilized in our study encompass SmoothNLP and Wealth-alpaca~\cite{maia201818} dataset. SmoothNLP dataset is tailored for the task of news headline generation, focusing primarily on generating appropriate titles based on financial news content. As for the Wealth-alpaca dataset, we extract questions from it and subsequently task ChatGPT with generating 2-3 key terms as output for our instruction data construction.}


\paragraph{Question Answering} ~ {We use Wealth-alpaca~\cite{maia201818} dataset, which is a combination of Stanford's Alpaca~\footnote{\url{https://github.com/tatsu-lab/stanford_alpaca}} and FiQA~\footnote{\url{https://sites.google.com/view/fiqa/}} with another 1.3k pairs custom generated using GPT3.5.}

\begin{table*}
\centering
\begin{tabular}{lllr} \toprule
Dataset            & Major Task Type             & Minor Task Type              & \# Samples  \\ \midrule
FPB                & Sentiment Analysis     & Sentiment Analysis      & 18690                           \\
FIQA-SA            & Sentiment Analysis     & Sentiment Analysis      & -          \\
FNSC               & Sentiment Analysis     & Sentiment Analysis      & -           \\
CCKS-NEC-2022       & Imformation Extraction & Causality Extraction    & 7499                            \\
SmoothNLP IEE      & Imformation Extraction & Event Extraction        & 3256                            \\
SmoothNLP NHG      & Text Generation        & Text Generation         & 4642                            \\
CCKS2022-event     & Text Classification & Event Type Classification   & 3578                            \\
Minds14            & Text Classification & Intent Prediction   & 59143                           \\
Financial Report   & Imformation Extraction & Entity Extraction       & 61705                           \\
OpenKG             & Imformation Extraction & Entity Extraction       & 7672                            \\
OpenKG             & Imformation Extraction & Entity Extraction       & 67921                           \\
FDDC2018           & Translation            & Terminology Translation & 333                             \\
Wealth-alpaca-lora & Text Generation        & Keyword Generation      & 41825                           \\
\bottomrule                         
\end{tabular}
\caption{Data statistics of our financial NLP datasets.}
\label{tab:nlp_datasets}
\end{table*}


\section{Details of FinCUGE evaluation benchmark}
\label{sec:bbt6}

In this section, we introduce the six financial NLP datasets involved in the FinCUGE evaluation benchmark in detail. These datasets are strictly non-overlapping with the financial NLP dataset used to build DISC-FinLLM-SFT in the previous section, while their task types are partially the same, such as financial sentiment analysis.  

\paragraph{FinFE} ~ {A financial social media text sentiment classification dataset. Given financial social media text, the model needs to classify the
sentiment of the text into negative-neutral-positive categories, with evaluation measured by accuracy. The training set contains 8,000 articles, the validation set contains 1,000 articles, and the test set contains 1,000 articles.}

\paragraph{FinQA} ~ {A financial news announcement event question-answering dataset, derived from the DuEE-fin~\cite{10.1007/978-3-031-17120-8_14} dataset. Given financial news or announcement text and a question related to an event mentioned in the text, the model needs to generate an answer to the question based on the text, with evaluation measured by F1-Score. The training set contains 16,000 articles, the validation set contains 2,000 articles, and the test set contains 2,000 articles.}

\paragraph{FinCQA} ~ {A financial causal event extraction dataset. Given financial news articles and a question related to a causal event mentioned in the text, the model needs to generate an answer to the question based on the text, with evaluation measured by F1-Score. The training set contains 21,965 articles, the validation set contains 2,741 articles, and the test set contains 2,745 articles.}

\paragraph{FinNA} ~ {A financial news summarization dataset. Given financial news articles, the model needs to generate a summary, with evaluation measured by Rouge. The training set contains 24,000 articles, the validation set contains 3,000 articles, and the test set contains 3,000 articles.}

\paragraph{FinRE} ~ {A financial news relation extraction dataset. Given financial news articles and head-tail entity pairs, the model needs to classify the relation between entity pairs into up to 44 categories, including the null relation, with evaluation measured by F1-Score. The training set contains 7,454 articles, the validation set contains 1,489 articles, and the test set contains 3,727 articles.}

\paragraph{FinESE} ~ {A financial news dataset. Given financial news articles, the model needs to extract the subjects of a specific event type, with evaluation measured by F1-Score. The training set contains 14,252 articles, the validation set contains 1,781 articles, and the test set contains 1,782 articles.}

It is actually difficult for us to be completely consistent with standard instruction tuning~\cite{wei2021finetuned}, that is, to evaluate on completely unseen task types. This is because the number of financial NLP datasets is relatively small, especially the high-quality Chinese financial NLP datasets, which to some extent hinders us from testing the zero-shot generalization ability of financial LLM on NLP tasks.












\end{document}